%% file: bmvc_final.tex
\documentclass{bmvc2k}
\usepackage{booktabs}
\usepackage{amssymb}
\usepackage{bbding}
\usepackage{hyperref}
\hypersetup{
    colorlinks=true,
    linkcolor=blue,
    filecolor=blue,      
    urlcolor=blue,
    citecolor=cyan,
}

\title{Cascaded Sparse Feature Propagation Network for Interactive Segmentation}

\addauthor{Chuyu Zhang*}{zhangchy2@shanghaitech.edu.cn}{1,3}
\addauthor{Chuanyang Hu*}{huchy3@shanghaitech.edu.cn}{1}
\addauthor{Hui Ren}{renhui@shanghaitech.edu.cn}{1}
\addauthor{Yongfei Liu}{liuyf3@shanghaitech.edu.cn}{1}
\addauthor{Xuming He}{hexm@shanghaitech.edu.cn}{1,2}

\addinstitution{
ShanghaiTech University \\
Shanghai, China
}
\addinstitution{
Shanghai Engineering \\
Research Center of \\
Intelligent Vision and Imaging \\
Shanghai, China
}

\addinstitution{
Lingang Laboratory \\
Shanghai, China
}

\runninghead{Zhang et al.}{Cascaded Sparse Feature Propagation Network}

\begin{document}

\maketitle
\def\thefootnote{*}\footnotetext{These authors contributed equally to this work.}
\input{sections/abstract.tex}

\input{sections/intro.tex}
\input{sections/related_work.tex}
\input{sections/method.tex}
\input{sections/experiments.tex}

\input{sections/conclusion.tex}

\bibliography{bmvc_final}

\end{document}

%% file: sections/abstract.tex
\begin{abstract}
We aim to tackle the problem of point-based interactive segmentation, in which the key challenge is to propagate the user-provided annotations to unlabeled regions efficiently. Existing methods tackle this challenge by utilizing computationally expensive fully connected graphs or transformer architectures that sacrifice important fine-grained information required for accurate segmentation. To overcome these limitations, we propose a cascade sparse feature propagation network that learns a click-augmented feature representation for propagating user-provided information to unlabeled regions. The sparse design of our network enables efficient information propagation on high-resolution features, resulting in more detailed object segmentation. We validate the effectiveness of our method through comprehensive experiments on various benchmarks, and the results demonstrate the superior performance of our approach. Code is available at \href{https://github.com/kleinzcy/CSFPN}{https://github.com/kleinzcy/CSFPN}.
\end{abstract}


%% file: sections/intro.tex
\section{Introduction}
Interactive image segmentation plays a vital role in a broad range of human-in-the-loop vision tasks, such as image editing~\cite{cheng2010repfinder}, medical image analysis~\cite{wang2018deepigeos} and dense image annotation~\cite{sofiiuk2020f}. There has been a long history of interactive segmentation in vision literature, in which a variety of interaction strategies have been explored, including points~\cite{xu2016deep,sofiiuk2020f}, scribbles~\cite{bai2014error,agustsson2019interactive}, and bounding boxes~\cite{rother2004grabcut}. In this work, we mainly focus on the point-based interactive segmentation that only provides point clicks to indicate foreground or background on an image, which typically requires less effort from human annotators.  
Unlike semantic segmentation, which only takes the input image, point-based interactive segmentation takes both image and extra human point annotation based on the segmentation map as input. Therefore, the main challenge of interactive segmentation is to propagate human-provided feedback information to unlabeled pixels effectively.

Most methods only utilize point clicks in the network input, where they encode human input as a Gaussian map and then utilize stacked convolutions to propagate user annotation in an implicit manner~\cite{xu2016deep,lin2020interactive}. This strategy has a limited capacity to capture long-range dependency, often leading to incomplete foreground masks. 
To tackle this, CDNet~\cite{chen2021conditional} adopt fully-connected graph networks~\cite{wang2018non} to facilitate information propagation at both global and local levels. Nonetheless, the fully-connected graph network can only cope with relatively low-resolution feature maps due to its high-computation complexity, which can result in inaccurate boundaries. Recently, ViT~\cite{dosovitskiy2020image} has been introduced into interactive segmentation, such as SAM~\cite{kirillov2023segment}. Although SAM can capture long-range dependency, it is limited by the resolution of ViT and hard to handle fine-grained segmentation. 

To capture long-range dependency and handle fine-grained segmentation, we propose a novel cascaded sparse feature propagation network for interactive image segmentation. Our main idea is to \textit{learn a click-augmented feature representation} based on a cascaded sparse graph neural network (GNN), which allows efficient long-range information propagation at a high spatial resolution, thus enabling us to generate more accurate segmentation. To this end, we introduce a new sparse graph network to propagate the user click information to the unlabeled region in a non-local yet efficient manner. 

Specifically, we first select a set of high-level feature representations corresponding to human-provided points as the source information, which is clean and informative, and propagate the selected source information to the unlabeled region by a sparse graph neural network. Such a sparse graph design is effective in propagating human-provided information but is unable to preserve high-resolution information, which is vitally important for interactive segmentation. Therefore, to preserve more detailed information, we propose another high-resolution sparse feature propagation network, which fuses high and low-level information and simultaneously propagates the fused information to the target region. Thanks to the sparsity of our design, our graph neural network can propagate information on feature maps of 1/2 image height and width, which is infeasible for fully connected graph networks. Such a cascaded network of two sparse GNNs is capable of computing a set of click-augmented feature representations at high spatial resolution in linear complexity, achieving high efficiency in propagating information and preserving more detailed information for foreground mask prediction. What's more, we adopt a global-to-local strategy to zoom-into human-interested regions for more accurate and efficient segmentation.

We conduct extensive experiments on six datasets, including GrabCut, Berkeley, DAVIS, COCO, and SBD, with a detailed ablation study. The results demonstrate the effectiveness of our method, which achieves the state-of-the-art performance.
Our contributions are summarized as the following:
\begin{itemize}
\item We propose a cascaded sparse feature propagation network for interactive image segmentation, which is capable of capturing long-range dependency on both low- and high-resolution feature maps and generating more accurate object boundaries. 
\item We develop a global-to-local strategy, which dynamically zooms into human-interested regions and provides more high-resolution information.
\item Our method achieves state-of-the-art results on most public benchmarks, demonstrating the effectiveness of our design. 
\end{itemize}

%% file: sections/related_work.tex
\section{Related works}
\paragraph*{Interactive segmentation:}
Interactive image segmentation has attracted much attention in computer vision research, and a variety of interaction strategies have been developed based on bounding boxes, scribbles, or points. While the bounding-box-based methods~\cite{rother2004grabcut,zhang2020interactive,wu2014milcut} can localize the target object quickly, and the scribble-based methods~\cite{grady2006random,bai2014error,gulshan2010geodesic,agustsson2019interactive} provide richer user-input cues, they often involve more user interactions. By contrast, the point-based, where a user provides points to indicate foregrounds or backgrounds on the image, requires less effort from human annotators~\cite{xu2016deep,sofiiuk2020f,sofiiuk2021reviving,chen2021conditional,benenson2019large,Hao_2021_ICCV}. Consequently, we mainly focus on the point-based methods in the discussion below.

Many point-based works have emerged since \cite{xu2016deep} first proposed a CNN-based method, and they can be largely grouped into two categories. While one trend focuses on annotating object boundaries~\cite{castrejon2017annotating,le2018interactive,acuna2018efficient,ling2019fast}, most deep learning methods perform region-based segmentation, aiming to leverage user clicks in each interactive step more efficiently. In particular, \cite{liew2017regional} attempts to refine local regions based on pairs of positive and negative clicks.  \cite{majumder2019content} generates a content-aware guidance map for exploiting the hierarchical structural information in the image. \cite{lin2020interactive} argues that the first click is more important than others and designs a first-click attention mechanism. \cite{Hao_2021_ICCV} improves the usage of interactive information from user clicks with the edge-guided flow. To better preserve detailed information, \cite{Chen_2022_CVPR,Lin_2022_CVPR} propose a local refinement strategy. To better adapt to test cases, \cite{jang2019interactive,sofiiuk2020f} develop a backpropagating refinement scheme to correct the mislabeled user clicks in the test time. 

In spite of their promising performances, existing approaches in the field typically rely on human-provided information in the input space. These methods encode positive and negative points provided by humans as Gaussian maps and expect the convolutional neural network (CNN) to propagate this information to unlabeled points. Additionally, several recent approaches~\cite{kirillov2023segment,liu2022simpleclick} have introduced transformer-based methods that effectively propagate user-provided information. However, these methods are limited in their resolution due to the constraints of the vision transformer architecture.

In contrast to these approaches, we propose a cascaded sparse feature propagation network that leverages human-provided information on both low- and high-resolution feature maps. This allows for more effective utilization of the human-provided points. Furthermore, our approach adopts a global-to-local strategy, which aims to exclude the distraction from background regions and accurately localize the regions of interest identified by humans. Our strategy is simple and complementary to previous methods that focused on local region refinement.


\paragraph{Graph neural network:}
Graph neural networks ~\cite{gori2005new,scarselli2008graph} have been widely adopted to capture long-range dependencies, and there exists a large body of literature on this research topic~\cite{wu2020comprehensive}. However, only a few works utilize GNNs in the task of interactive segmentation. For instance, based on the non-local networks~\cite{wang2018non}, CDNet~\cite{chen2021conditional} proposes a conditional diffusion network for interactive segmentation, which performs non-local feature propagation on the global convolutional features and local-level pixels using color similarity.
This mixed strategy tends to suffer from inaccurate foreground boundaries due to the low-resolution deep feature map and/or the noisy graph affinity estimated based on previous foreground masks or color similarity. 
In contrast, we propose a simple sparse attention-based non-local graph network to propagate the click information,  which can be applied to a high-resolution feature map and generate more accurate masks.

%% file: sections/method.tex
\section{Method}
\begin{figure*}[!t]
	\centering
	\includegraphics[scale=0.55]{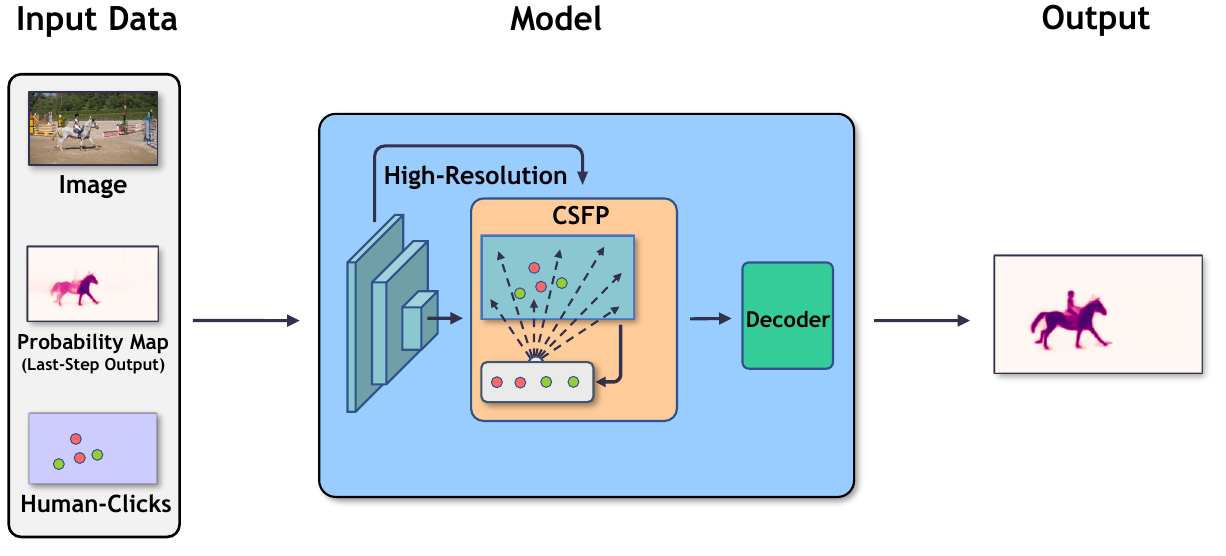}
	\caption{The overall architecture of cascaded sparse feature propagation network (CSFP). We take as input an image, the previous step probability map and a set of human clicks. In the feature space, to fully utilize human-provided information, we select human-provided click information and propagate that information by a cascaded of sparse graphs.
}
	\label{fig:Overview}
\end{figure*}
\subsection{Method Overview}\label{overview}
Interactive segmentation aims to correctly infer the region of the user's interest and segment the target object with as few clicks as possible. 
This sequential estimation task is typically converted into a series of foreground segmentation problems, each of which aims to output a foreground mask as accurately as possible.

In this work, we focus on the key aspects of click-guided foreground mask prediction to efficiently propagate the click labels to unlabeled regions. To this end, we propose a novel cascaded sparse feature propagation module, which performs information propagation on both low- and high-resolution feature maps as shown in Fig.\ref{fig:Overview}. Our model employs the newly designed sparse GNNs to facilitate the information propagation of the user's inputs. In addition, we develop a global-to-local strategy to zoom into high-resolution human-interested regions for more efficient segmentation. Below, we introduce the details of our model design.


\subsection{Cascaded Sparse Feature Propagation}\label{prop}
We aim to propagate the user-provided information to unlabeled regions in the input images. Due to the sparsity of user clicks, this typically requires modeling long-range feature relations across the image plane. To achieve efficient information propagation, we introduce a new graph neural network module, which augments a base CNN segmentation network for predicting the foreground mask. In contrast to previous non-local design~\cite{wang2018non,chen2021conditional}, our graph network module is built on a sparse graph topology, which enables us to compute a user-input-aware representation on a high-resolution feature map and hence produces detailed foreground segmentation with accurate boundaries.   

Specifically, we first use a CNN-based segmentation network to compute a stack of feature maps from which a higher-level and a lower-level map are selected. The higher-level feature map, denoted as $\mathcal{F}$, typically encodes more semantics but has a low resolution, while the lower-level map, denoted as $\mathcal{F}^h$, has a high resolution and preserves more object boundary cues. The size of $\mathcal{F}$ and $\mathcal{F}^h$ are 1/16 and 1/4 of the original image size. 
In order to better exploit both the higher- and lower-level features, we employ a cascaded design: a sparse graph network first augments the higher-level feature map with the user-click features, which is further integrated with the lower-level features in a high-res sparse graph network at the second stage. 
Below we describe the details of those two graph networks in turn.  

\paragraph{Sparse Graph sub-Module (SGM):} 
Our sparse graph submodule performs feature propagation on the higher-level feature maps to disseminate the user-click information to all features. 
Formally, we represent the higher-level feature map $\mathcal{F}$ as $ \{f_{n}\}_{n=1}^{H_l\times W_l}$, where $H_l$ and $W_l$ are the height and width of the low-res feature map respectively. At each location, $f_{n} \in \mathbb{R}^c$ is a $c$-channel feature vector, and $n$ is the spatial location index. 
To build the sparse graph, we select the feature vectors at the location of user clicks in $U$, denoted as $\mathcal{F}_u = \{f_{u_i}\}_{i=1}^{M}$, and connect them to each location on the feature map.  

\begin{figure*}[!t]
	\centering
	\includegraphics[scale=0.35]{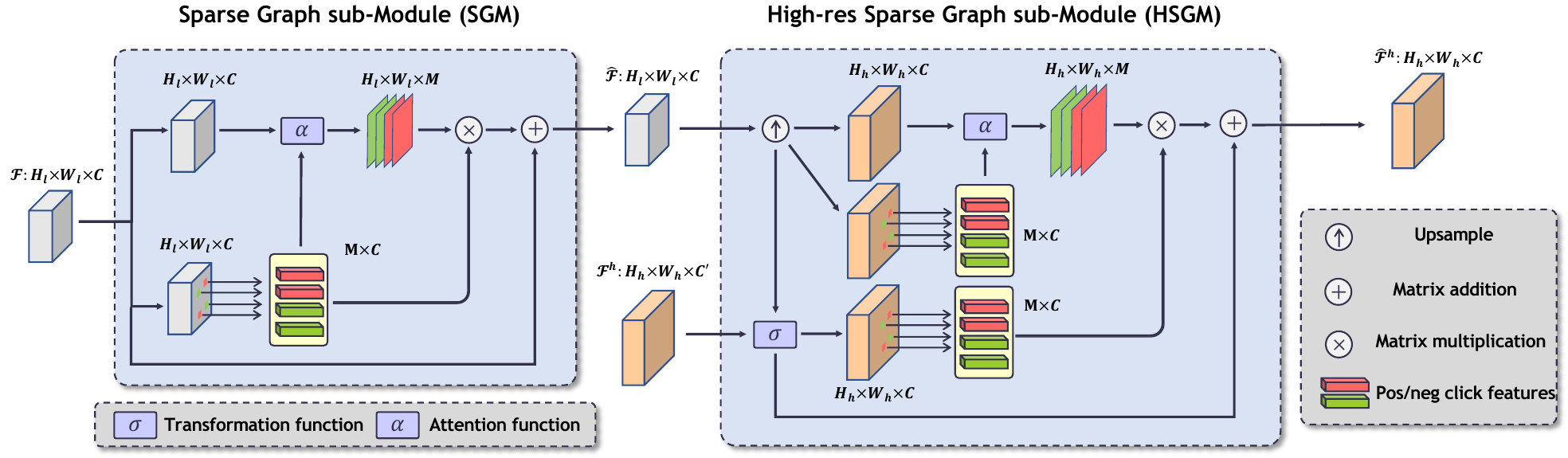}
	\caption{The structure of sparse graph sub-module(SGM) and high-resolution sparse graph sub-module(HSGM). $H$ and $W$ represent the height and width of the feature map. $C$ and $C^{'}$ represent the channel of higher-level and lower-level feature maps, respectively. $M$ is the number of user clicks. For simplicity, we ignore the reshape operation.}
	\label{fig:labelpropagation}
\end{figure*}

Given the graph, we perform feature augmentation by passing messages from the click nodes $\mathcal{F}_u$ to each feature location as follows (See Fig.~\ref{fig:labelpropagation} for illustration):
\begin{align}
	&\hat{f}_n = {f}_{n} + \sum_{j=1}^{M}\alpha(f_n, f_{u_j})W_c^\intercal f_{u_j}, \; \forall f_n \in \mathcal{F} , \\
	&\alpha(f_n, f_{u_j})=e^{\theta (f_n)^\intercal \Phi(f_{u_j})}/Z_n(\mathcal{F}),
\end{align}
where $\hat{f}_{n}$ is the updated feature, $W_c\in \mathbb{R}^{c\times c}$ is a weight matrix for feature transform, and  $\alpha$ denotes an attention function in which $\theta$ and $\phi$ are linear transforms and $Z_n(\mathcal{F})$ is the normalization factor. 
Intuitively, the augmentation moves all the features towards the click-annotated representations, which reduces in-class variation and improves foreground prediction. However, these augmented features, denoted as $\hat{\mathcal{F}}$, are built on the low-res feature map $\mathcal{F}$, which tends to produce coarse segmentation masks. To remedy this, we introduce a second graph network to refine them as below. 

\paragraph{High-res Sparse Graph sub-Module (HSGM):}
The second graph network generates a click-augmented feature representation with a high spatial resolution. To this end, we integrate the first-stage output $\hat{\mathcal{F}}$ with the lower-level feature map $\mathcal{F}^h$, followed by another pass of click-to-feature propagation. 
Specifically, we denote the lower-level feature map as 
$\{f_n^h\}_{n = 1}^{H_h\times W_h}$ where $H_h$ and $W_h$ are the height and width of the high-res feature map, respectively. Similar to the SGM, we select the click-annotated feature, represented as $\mathcal{F}_u^{h} = \{f_{u_j}^h\}_{j=1}^{M}$, and link them to every feature location on $\mathcal{F}^h$.

Given the high-res graph, we first upsample the previous output $\hat{\mathcal{F}}$ so that it has the same spatial dimension as the lower-level feature map $\mathcal{F}^h$. We then perform feature integration and click-aware augmentation by a message passing as shown in Fig.~\ref{fig:labelpropagation}. Formally, denoting the upsampled feature as $\{\hat{g}_n\}_{n = 1}^{H_h\times W_h}$, we update the high-res feature representation as follows,
\begin{align}
	\hat{f}^{h}_n = \sigma(f^h_{n}\oplus\hat{g}_{n}) + \sum_{j=1}^{M}\alpha(\hat{g}_{n}, \hat{g}_{u_j})W_f^\intercal \sigma(f^h_{u_j}\oplus \hat{g}_{u_j}),
\end{align}
where $\hat{f}^{h}_n$ denotes the high-res augmented feature, $\oplus$ indicates feature concatenation, $\sigma(\cdot)$ is a transformation function and $W_f$ is a weight matrix for feature transform. Note that we use the higher-level features $\{\hat{g}_n\}$ to compute the attention weights so that the information propagation is less susceptible to variations in the lower-level feature $\mathcal{F}^h$. Moreover, our sparse graphs only perform message passing from $M$ selected nodes to $N$ feature nodes, which can be computed efficiently with a complexity $\mathcal{O}(MN)$ where $M \ll N$ in the interaction. 

Given the high-res augmented features $\{\hat{f}_n^h\}_{n = 1}^{H_h\times W_h}$, we finally generate the foreground probability map by applying a two-layer conv-block followed by the Sigmoid function. 

\subsection{Model Training}\label{train}
To train our deep network for interactive segmentation, we first utilize a simulation process to generate a set of image-click pairs from a foreground segmentation dataset as in~\cite{sofiiuk2021reviving}, which assumes the user clicks on the center of maximum error regions. Given the training dataset, 
following \cite{sofiiuk2020f}, we employ the Normalized Focal Loss (NFL)~\cite{sofiiuk2019adaptis} as training objectives. 

\subsection{Global to local strategy}
In interactive segmentation, users have specific areas of interest within an image. Performing segmentation on the entire image is typically less effective than segmenting the user-interested region due to the limited resolution. To achieve more accurate segmentation results in areas of user interest, we propose a global-to-local strategy in the interactive process.

Our global-to-local strategy consists of three steps for each foreground segmentation task. We first perform object segmentation on the entire image by utilizing the segmentation map from the previous interaction round and the human correction click in the current interaction. Subsequently, based on the segmentation map obtained in the first
step, we determine the bounding box of the target and then expand it by a margin, such as $40\%$ of its original size. Finally, we focus on the region defined by the expanded box and perform target segmentation within this zoomed-in area. By adopting this global-to-local strategy, our model can efficiently segment the areas that users care about and enhance the segmentation quality.



%% file: sections/experiments.tex
\vspace{-3mm}
\section{Experiments}

In this section, we first describe the experiment setting and implementation details, 
then compare our model with existing works, followed by the ablation study to validate each component. 
Finally, we show some qualitative results to demonstrate the model's efficacy.


\vspace{-2mm}
\subsection{Evaluation and Implementation Details}

\paragraph{Datasets:} We evaluate our method on a wide range of datasets, including GrabCut, Berkeley, DAVIS, COCO and SBD, with the standard evaluation protocol. COCO is split into COCO$^s$ and COCO$^u$ according to whether their object classes are in PASCAL VOC or not.
%







\vspace{-2mm}
\paragraph{\textbf{Metric:}} To mimic the real user clicks in evaluation, we follow \cite{xu2016deep} to click the center of the maximum error region to correct the output mask in each interaction. The interaction process will terminate when the IoU between prediction and ground truth mask exceeds threshold $\tau$, or reaches the maximum number of clicks. 
In this paper, we typically set $\tau$ as 85\% or 90\%, and set a maximum number of clicks to 20 as in previous works.
The number of clicks (NoC) and the number of failures (NoF) to meet the termination conditions is used as the evaluation metric. For example, NoC@90 means the average number of clicks for the test set is needed to reach 90\% IoU under 20 maximum clicks, and NoF@90 means the number of failures case that does not reach 90\% IoU with 20 maximum clicks.

\vspace{-2mm}
\paragraph{\textbf{Implementation Details:}} We adopt DeeplabV3+\cite{Chen_2018_ECCV} as our base network.
Besides, we follow \cite{sofiiuk2021reviving} to utilize Conv1S
to fuse click maps and foreground estimation, then sum the fused feature with image features at the output of the first convolutional block.
During training, we follow the same iterative sampling strategy in \cite{sofiiuk2021reviving} to 
generate positive and negative clicks to alleviate the gap between the training and inference stages.
For clicks encoding, we adopt the disk encoding strategy proposed in \cite{benenson2019large} with a fixed radius of 5.
Our network is trained for 120 epochs using the SBD training set and COCO+LVIS dataset. We apply common data augmentation techniques, such as random cropping and scaling. The Adam optimizer is used for optimization with a learning rate of 5e-4. The learning rate is reduced by a factor of 10 at the 100th and 115th epochs. We use a batch size of 28, and the augmented images are resized to $320\times 480$.

\begin{table*}[t]
    \centering
    \caption{\small{Comparison with SOTA. The bold means the best results across different backbones, and the underline means the best results under the same backbone. * means that we implement the results. $^\dagger$ denotes the model is trained on COCO+LVIS, while others are trained on the SBD dataset. - denotes the results are not available. The lower is the better.}}
    \label{sota_Table}
    \renewcommand\arraystretch{1.3} 
    \resizebox{1.0\textwidth}{!}{
    \begin{tabular}{ll|cccccc}
    \toprule[1pt]
    \multicolumn{2}{c|}{\textbf{Method}}        &\textbf{GrabCut} &\textbf{Berkeley}  &\textbf{COCO$^s$} &\textbf{COCO$^u$} & \textbf{DAVIS} & \textbf{SBD} \\ 
    &                                           &\textbf{NoC@90}  & \textbf{NoC@90}  & \textbf{NoC@85} & \textbf{NoC@85} & \textbf{NoC@85 / 90}  & \textbf{NoC@85 / 90} \\
    \midrule\midrule

    CDNet\cite{chen2021conditional}  \scriptsize{\texttt{ICCV21}}                          & ResNet-50       & 2.64 & 3.69   & -    & -   & 5.17~/~6.66 &  4.37~/~7.87  \\
    FocusCut\cite{Lin_2022_CVPR} \scriptsize{\texttt{CVPR22}}  & ResNet-50 & \underline{1.78} & 3.44 & - & - & 5.00~/~6.38 & 3.62~/~5.66 \\
    Ours                                                     & ResNet-50       & 2.31 & \underline{3.35}   & \underline{2.46} & \underline{3.69}  & \underline{4.52}~/~\underline{5.83}  & \underline{3.08}~/~\underline{4.98}  \\ 
    \midrule

    IS+SA\cite{zhang2020interactive}  \scriptsize{\texttt{ECCV20}}                    & ResNet-101      & 3.07 & 4.94      & 4.08 & 5.01 & 5.16~/~-    &    -~/~-   \\
    FCA\cite{lin2020interactive} \scriptsize{\texttt{CVPR20}}                              & ResNet-101      & 2.14 & 4.19   & 4.45 & 5.33 &  -~/~7.90   &    -~/~-   \\
    f-BRS-B\cite{sofiiuk2020f}  \scriptsize{\texttt{CVPR20}}                              & ResNet-101      & 2.72 & 4.57   & -    & -    & 5.04~/~7.41 & 4.81~/~7.73  \\ 
    RITM*\cite{sofiiuk2021reviving} \scriptsize{\texttt{ARXIV}}                       & ResNet-101     & 2.31 & 3.50   & 2.54 & 3.61 & 5.09~/~6.78 &   3.33~/~5.33 \\
    FocusCut\cite{Lin_2022_CVPR} \scriptsize{\texttt{CVPR22}} & ResNet-101 & \underline{1.64} & \underline{3.01} & - & - & 4.85~/~6.22 & 3.40~/~5.31 \\
    \textbf{Ours}                                                     & ResNet-101      & 2.15 & 3.20 & \underline{2.27} & \underline{3.50} & \underline{4.51} ~/~ \underline{5.80} &  \underline{\textbf{2.98}}~/~\underline{\textbf{4.83}}  \\ \midrule
    
    RITM\cite{sofiiuk2021reviving} \scriptsize{\texttt{ARXIV}}                  & HRNet-18        & 2.04 & 3.22   & 2.40    & 3.61   & 4.94~/~6.71 &  3.39~/~5.43  \\
    PseudoClick\cite{liu2022pseudoclick} \scriptsize{\texttt{ECCV22}} & HRNet-18 & 2.04 & 3.23 & - & - & 4.81~/~6.57 & -~/~5.40 \\
    \textbf{Ours}                                                & HRNet-18        & \underline{1.75} & \underline{2.86} & \underline{2.23} & \underline{3.00} & \underline{4.68}~/~ \underline{6.01} &  \underline{3.33}~/~\underline{5.25} \\ \midrule
    SAM\cite{kirillov2023segment} \scriptsize{\texttt{ARXIV}}                 & ViT-H        & 1.84 & \textbf{2.09}   & \multicolumn{2}{c}{5.16}    & 4.21~/~5.32 & 5.23~/~8.50  \\ \midrule
    FocalClick$^\dagger$\cite{Chen_2022_CVPR} \scriptsize{\texttt{CVPR22}} & HRNet-32 & 1.80 & 2.36 & - & - & \underline{\textbf{4.01}}~/~5.39 & 4.24~/~6.51 \\
    RITM$^\dagger$\cite{sofiiuk2021reviving} \scriptsize{\texttt{ARXIV}}                 & HRNet-18        & \underline{\textbf{1.54}} & 2.26   & -    & -    & 4.36~/~5.74 &  3.80~/~6.06  \\
    \textbf{Ours$^\dagger$}                                                 & HRNet-18        & 1.68 & \underline{2.12} & \multicolumn{2}{c}{\textbf{\underline{2.17}}} & 4.03~/~ \underline{\textbf{5.22}} &  \underline{3.67}~/~\underline{5.91} \\    
    \bottomrule[1pt]
    \end{tabular}}

\end{table*}

\subsection{Quantitative Results}

\paragraph{\textbf{Main results:}} 
As shown in Tab.\ref{sota_Table}, we compare our method with previous approaches over a wide range of benchmarks. When trained on the SBD dataset, our method achieves the strongest performance over different datasets, outperforming all existing approaches under the same backbone. We achieve the best results under HRNet-18 backbone, which only requires average \textbf{2.86} clicks to reach 90\% IoU on Berkeley, \textbf{2.23} clicks to reach 85\% IoU on COCO$^s$ datasets. The result on COCO$^u$ shows that our framework can generalize to unseen classes greatly. Specifically, for the challenging SBD and fine-grained DAVIS dataset, we achieve average \textbf{5.8} and \textbf{4.83} NoC@90 respectively on the ResNet-101 backbone, improve \textbf{0.42} and \textbf{0.48} compared with FocusCut. 


Moreover, we train our model on the COCO+LVIS dataset. The results in the last three lines show that we still achieve a great improvement on DAVIS. Meanwhile, like RITM, we also observe a drop in SBD datasets compared to those trained on SBD datasets. Furthermore, our method achieves similar or better results than SAM which is trained on massive data. Those results show the superiority of our method.

\begin{figure}[!t]
    \centering
    \includegraphics[width=0.6\textwidth]{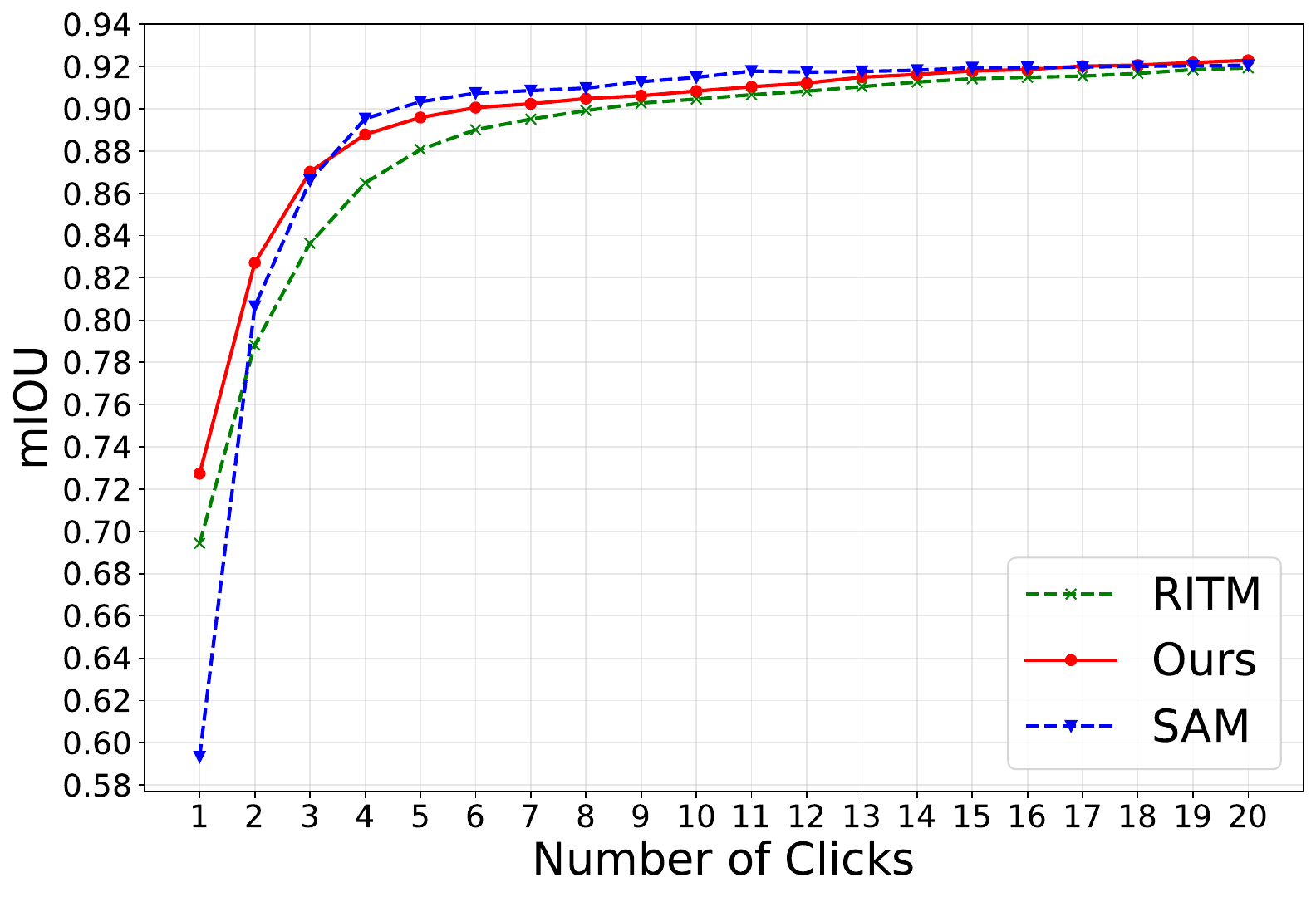}
    \caption{\small{The improvement of mIOU across different numbers of clicks k on DAVIS.}}
    \label{fig:foobar}
\end{figure}

\vspace{-1mm}
\paragraph{\textbf{IOU@k Analysis:}} We analyse the performance varying over the number of clicks k. As shown in the Fig.\ref{fig:foobar}, our method is consistently better than RITM and about 2\% higher in mIOU for the first five clicks. Moreover, our method outperforms SAM in the first three clicks, and achieves similar results with more than 14 clicks. The results validate the effectiveness of our method.


\paragraph{\textbf{Results with maximum 100 clicks:}} 
We report NoC with the maximum number of clicks limited to 100 on the DAVIS dataset with ResNet-50 backbone. Tab.\ref{tb:a100c} shows that our method improves significantly on NoC@90. Moreover, we improve 0.91 on NoC$_{100}$@90 metric compared with CDNet. This is mainly because our sparse graph can propagate information on high-resolution feature maps which preserve more fine-grained feature information, while CDNet lacks such modeling capacity.

\begin{table}[t]
\centering
\begin{minipage}[t]{0.48\textwidth}
        \centering
             \caption{\small{Analysis of 100 clicks.}}
     \label{tb:a100c}
        \renewcommand\arraystretch{1.2}
        \resizebox{0.9\textwidth}{!}{
        \begin{tabular}{c|ccc}
            \toprule[1pt]
            \textbf{Method}   & \textbf{NoF@90} & \textbf{NoC$_{100}$@90} & \textbf{NoF$_{100}$@90} \\ \midrule\midrule
            f-BRS\cite{sofiiuk2020f}              & 78            & 20.70          & 50             \\
            CDNet\cite{chen2021conditional}              & 65            & 18.59          & 48             \\
            FocusCut\cite{Lin_2022_CVPR}              & 57            & \textbf{17.42}          & \textbf{43}             \\
            Ours               & \textbf{56}            & 17.68          & 46       \\ \bottomrule[1pt]       
        \end{tabular}}
\end{minipage} \hfill
\begin{minipage}[t]{0.5\textwidth}
    \centering
         \caption{Component Analysis.}
 \label{CSFP-verification}
    \renewcommand\arraystretch{1.2}
    \resizebox{0.9\textwidth}{!}{	
    \begin{tabular}{c|cccc|cc}
        \toprule[1pt]
        \textbf{\#} & \textbf{BS}  & \textbf{SGM}  & \textbf{HSGM} & \textbf{G2L} & \textbf{DAVIS} & \textbf{SBD} \\ \midrule
        1          & $\checkmark$  & -             & -   & -          & 6.85           & 5.67          \\
        2          & $\checkmark$  & $\checkmark$  & -   & -          & 6.62           & 5.57          \\
        3          & $\checkmark$  & $\checkmark$  & $\checkmark$ & -  & 6.39  & 5.36 \\ 
        4          & $\checkmark$  & $\checkmark$  & $\checkmark$ & $\checkmark$  & \textbf{6.05}      & \textbf{5.13} \\\bottomrule[1pt]
    \end{tabular}
        }
\end{minipage}
\vspace{-1em}
\end{table}

\subsection{Ablation study}

\paragraph{\textbf{Effectiveness of each component:}}  We conduct incremental ablation experiments to study the effectiveness 
of each component. As shown in Tab.\ref{CSFP-verification}, the introduction of \textit{Sparse Graph sub-Module (SGM)} brings 0.23 and 0.1 NoC@90 improvements on DAVIS and SBD, respectively. And the introduction of HSGM improves performance further, which means that feature propagation on high-resolution does help to preserve more precise boundary information. Moreover, our results boost significantly with global to local strategy (G2L).
\begin{table*}[!tp]
    \centering
     \caption{\small{Graph design analysis on DAVIS. 
 Low resolution denotes 1/4 feature map while high resolution represents 1/2 scale feature map.}}
 \label{graph-compare}
    \resizebox{0.8\textwidth}{!}{
    \begin{tabular}{c|cc|ccc}
        \toprule[1pt]
        \textbf{Method} & \textbf{Params(M)} & \textbf{Flops(G)}  & \textbf{NoC$_{20}@85$} & \textbf{NoC$_{20}@90$} \\ \midrule\midrule
        Baseline*       &  31.4         & 508.72                & 6.60           & 8.42         \\
        Baseline* + FDM~\cite{chen2021conditional}   & 31.42      & 513.82                & 5.40           & 7.64         \\
        Baseline* + FDM in both low \& high res.     & 31.44        & 1510.16                  & \XSolidBrush    & \XSolidBrush \\
        Baseline* + CSFP      & 31.5       & 531.42                  & \textbf{5.05}           & \textbf{7.17}        \\ \bottomrule[1pt]
        \end{tabular}}

\end{table*}

\begin{figure*}[!t]
    \centering
    \includegraphics[scale=0.3]{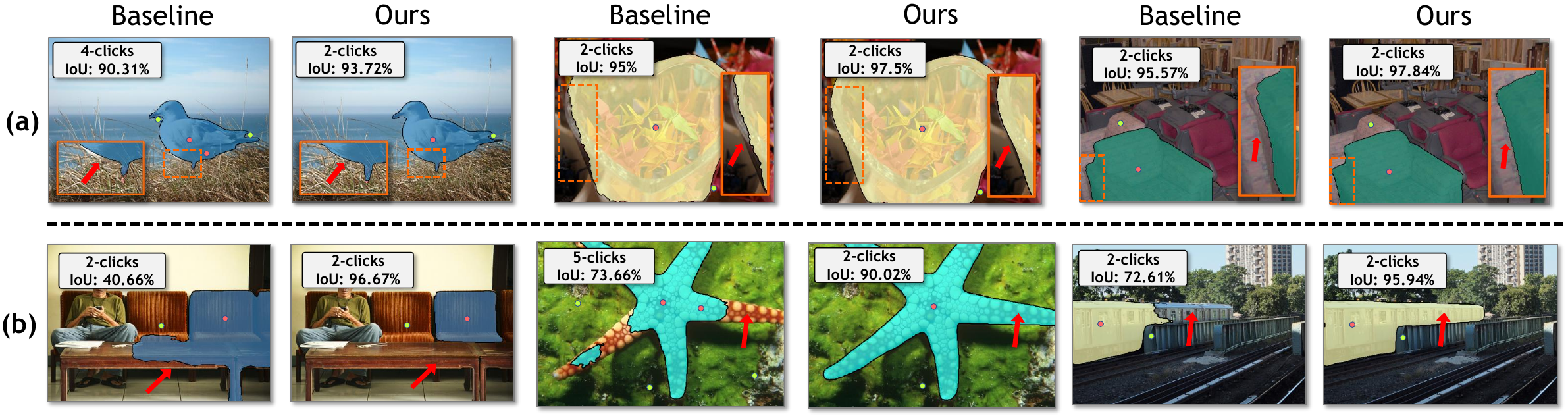}
    \caption{\small{\textbf{Visualization analysis:} The odd and even columns show the prediction result of the baseline and our method. 
Row (a) indicates that our method preserves more accurate boundary information, and Row (b) indicates that our method captures more 
    reliable long-range dependencies}}
    \label{vis-fig:2}
\end{figure*}

\paragraph{\textbf{Sparse Graph Analysis:}}
To validate the efficacy of CSFP design, we conduct more experiments to compare with feature diffusion graph module (FDM) in CDNet~\cite{chen2021conditional}, which is a fully-connected graph network. 
Since our method is easy to implement comparing to FDM, we immigrate our CSFP to CDNet’s baseline (named as Baseline*) and adopt the same training setup as CDNet. As in Tab~\ref{graph-compare}, we can observe that our `Baseline* + CSFP' achieves significant improvement in both metrics. It is worth noting that our advanced sparse graph design makes it feasible to conduct message propagation in both low \& high-resolution feature maps. 
In addition, when applying FDM in both low \& large scale feature maps, we find that the overall network consumes an amount of computation (1510.16 GFLOPs \textit{v.s.} 531.42 GFLOPs), which is unacceptable in real systems.

\vspace{-2mm}
\subsection{\textbf{Visualization analysis}}
\vspace{-1mm}
In Fig.\ref{vis-fig:2}(a), we observe that our method can obtain more precise boundaries even with fewer user clicks. This shows the effectiveness of the CSFP module and global-to-local strategy, which helps us preserve more accurate target region information. In Fig.\ref{vis-fig:2}(b), the results indicate that our sparse graph neural network can capture more reliable long-range dependencies and make the coverage of its prediction more complete.

%% file: sections/conclusion.tex
\section{Conclusion}
In this paper, we have developed a novel cascaded sparse feature propagation network for interactive segmentation. Our method is capable of effectively propagating user-provided sparse annotations to the entire input image. To achieve this, we introduce a cascaded sparse feature propagation module that selects user-provided information in the feature space and propagates the provided information to the whole image in high resolution. In addition, we propose a global-to-local strategy in the evaluation, which can accurately locate human-interested regions and zoom into this region, enabling our method to preserve more detailed information. Finally, we evaluated our method on several benchmarks, in which our approach outperforms the prior works by a sizable margin.
\section*{Acknowledgement}
This work was supported by Shanghai Science and Technology Program 21010502700, Shanghai Frontiers Science Center of Human-centered Artificial Intelligence and the MoE Key Lab of Intelligent Perception and Human-Machine Collaboration (ShanghaiTech University).

%% file: bmvc_final.bbl
\begin{thebibliography}{35}
\providecommand{\natexlab}[1]{#1}
\providecommand{\url}[1]{\texttt{#1}}
\expandafter\ifx\csname urlstyle\endcsname\relax
  \providecommand{\doi}[1]{doi: #1}\else
  \providecommand{\doi}{doi: \begingroup \urlstyle{rm}\Url}\fi

\bibitem[Acuna et~al.(2018)Acuna, Ling, Kar, and Fidler]{acuna2018efficient}
David Acuna, Huan Ling, Amlan Kar, and Sanja Fidler.
\newblock Efficient interactive annotation of segmentation datasets with
  polygon-rnn++.
\newblock In \emph{Proceedings of the IEEE conference on Computer Vision and
  Pattern Recognition}, pages 859--868, 2018.

\bibitem[Agustsson et~al.(2019)Agustsson, Uijlings, and
  Ferrari]{agustsson2019interactive}
Eirikur Agustsson, Jasper~RR Uijlings, and Vittorio Ferrari.
\newblock Interactive full image segmentation by considering all regions
  jointly.
\newblock In \emph{Proceedings of the IEEE/CVF Conference on Computer Vision
  and Pattern Recognition}, pages 11622--11631, 2019.

\bibitem[Bai and Wu(2014)]{bai2014error}
Junjie Bai and Xiaodong Wu.
\newblock Error-tolerant scribbles based interactive image segmentation.
\newblock In \emph{Proceedings of the IEEE Conference on Computer Vision and
  Pattern Recognition}, pages 392--399, 2014.

\bibitem[Benenson et~al.(2019)Benenson, Popov, and Ferrari]{benenson2019large}
Rodrigo Benenson, Stefan Popov, and Vittorio Ferrari.
\newblock Large-scale interactive object segmentation with human annotators.
\newblock In \emph{Proceedings of the IEEE/CVF Conference on Computer Vision
  and Pattern Recognition}, pages 11700--11709, 2019.

\bibitem[Castrejon et~al.(2017)Castrejon, Kundu, Urtasun, and
  Fidler]{castrejon2017annotating}
Lluis Castrejon, Kaustav Kundu, Raquel Urtasun, and Sanja Fidler.
\newblock Annotating object instances with a polygon-rnn.
\newblock In \emph{Proceedings of the IEEE conference on computer vision and
  pattern recognition}, pages 5230--5238, 2017.

\bibitem[Chen et~al.(2018)Chen, Zhu, Papandreou, Schroff, and
  Adam]{Chen_2018_ECCV}
Liang-Chieh Chen, Yukun Zhu, George Papandreou, Florian Schroff, and Hartwig
  Adam.
\newblock Encoder-decoder with atrous separable convolution for semantic image
  segmentation.
\newblock In \emph{Proceedings of the European Conference on Computer Vision
  (ECCV)}, September 2018.

\bibitem[Chen et~al.(2021)Chen, Zhao, Yu, Zhang, and Duan]{chen2021conditional}
Xi~Chen, Zhiyan Zhao, Feiwu Yu, Yilei Zhang, and Manni Duan.
\newblock Conditional diffusion for interactive segmentation.
\newblock In \emph{Proceedings of the IEEE/CVF International Conference on
  Computer Vision}, pages 7345--7354, 2021.

\bibitem[Chen et~al.(2022)Chen, Zhao, Zhang, Duan, Qi, and
  Zhao]{Chen_2022_CVPR}
Xi~Chen, Zhiyan Zhao, Yilei Zhang, Manni Duan, Donglian Qi, and Hengshuang
  Zhao.
\newblock Focalclick: Towards practical interactive image segmentation.
\newblock In \emph{Proceedings of the IEEE/CVF Conference on Computer Vision
  and Pattern Recognition (CVPR)}, pages 1300--1309, June 2022.

\bibitem[Cheng et~al.(2010)Cheng, Zhang, Mitra, Huang, and
  Hu]{cheng2010repfinder}
Ming-Ming Cheng, Fang-Lue Zhang, Niloy~J Mitra, Xiaolei Huang, and Shi-Min Hu.
\newblock Repfinder: finding approximately repeated scene elements for image
  editing.
\newblock \emph{ACM Transactions on Graphics (TOG)}, 29\penalty0 (4):\penalty0
  1--8, 2010.

\bibitem[Dosovitskiy et~al.(2020)Dosovitskiy, Beyer, Kolesnikov, Weissenborn,
  Zhai, Unterthiner, Dehghani, Minderer, Heigold, Gelly,
  et~al.]{dosovitskiy2020image}
Alexey Dosovitskiy, Lucas Beyer, Alexander Kolesnikov, Dirk Weissenborn,
  Xiaohua Zhai, Thomas Unterthiner, Mostafa Dehghani, Matthias Minderer, Georg
  Heigold, Sylvain Gelly, et~al.
\newblock An image is worth 16x16 words: Transformers for image recognition at
  scale.
\newblock \emph{arXiv preprint arXiv:2010.11929}, 2020.

\bibitem[Gori et~al.(2005)Gori, Monfardini, and Scarselli]{gori2005new}
Marco Gori, Gabriele Monfardini, and Franco Scarselli.
\newblock A new model for learning in graph domains.
\newblock In \emph{Proceedings. 2005 IEEE International Joint Conference on
  Neural Networks, 2005.}, volume~2, pages 729--734. IEEE, 2005.

\bibitem[Grady(2006)]{grady2006random}
Leo Grady.
\newblock Random walks for image segmentation.
\newblock \emph{IEEE transactions on pattern analysis and machine
  intelligence}, 28\penalty0 (11):\penalty0 1768--1783, 2006.

\bibitem[Gulshan et~al.(2010)Gulshan, Rother, Criminisi, Blake, and
  Zisserman]{gulshan2010geodesic}
Varun Gulshan, Carsten Rother, Antonio Criminisi, Andrew Blake, and Andrew
  Zisserman.
\newblock Geodesic star convexity for interactive image segmentation.
\newblock In \emph{2010 IEEE Computer Society Conference on Computer Vision and
  Pattern Recognition}, pages 3129--3136. IEEE, 2010.

\bibitem[Hao et~al.(2021)Hao, Liu, Wu, Han, Chen, Chen, Chu, Tang, Yu, Chen,
  and Lai]{Hao_2021_ICCV}
Yuying Hao, Yi~Liu, Zewu Wu, Lin Han, Yizhou Chen, Guowei Chen, Lutao Chu,
  Shiyu Tang, Zhiliang Yu, Zeyu Chen, and Baohua Lai.
\newblock Edgeflow: Achieving practical interactive segmentation with
  edge-guided flow.
\newblock In \emph{Proceedings of the IEEE/CVF International Conference on
  Computer Vision (ICCV) Workshops}, pages 1551--1560, October 2021.

\bibitem[Jang and Kim(2019)]{jang2019interactive}
Won-Dong Jang and Chang-Su Kim.
\newblock Interactive image segmentation via backpropagating refinement scheme.
\newblock In \emph{Proceedings of the IEEE/CVF Conference on Computer Vision
  and Pattern Recognition}, pages 5297--5306, 2019.

\bibitem[Kirillov et~al.(2023)Kirillov, Mintun, Ravi, Mao, Rolland, Gustafson,
  Xiao, Whitehead, Berg, Lo, et~al.]{kirillov2023segment}
Alexander Kirillov, Eric Mintun, Nikhila Ravi, Hanzi Mao, Chloe Rolland, Laura
  Gustafson, Tete Xiao, Spencer Whitehead, Alexander~C Berg, Wan-Yen Lo, et~al.
\newblock Segment anything.
\newblock \emph{arXiv preprint arXiv:2304.02643}, 2023.

\bibitem[Le et~al.(2018)Le, Mai, Price, Cohen, Jin, and Liu]{le2018interactive}
Hoang Le, Long Mai, Brian Price, Scott Cohen, Hailin Jin, and Feng Liu.
\newblock Interactive boundary prediction for object selection.
\newblock In \emph{Proceedings of the European Conference on Computer Vision
  (ECCV)}, pages 18--33, 2018.

\bibitem[Liew et~al.(2017)Liew, Wei, Xiong, Ong, and Feng]{liew2017regional}
JunHao Liew, Yunchao Wei, Wei Xiong, Sim-Heng Ong, and Jiashi Feng.
\newblock Regional interactive image segmentation networks.
\newblock In \emph{2017 IEEE international conference on computer vision
  (ICCV)}, pages 2746--2754. IEEE Computer Society, 2017.

\bibitem[Lin et~al.(2020)Lin, Zhang, Chen, Cheng, and Lu]{lin2020interactive}
Zheng Lin, Zhao Zhang, Lin-Zhuo Chen, Ming-Ming Cheng, and Shao-Ping Lu.
\newblock Interactive image segmentation with first click attention.
\newblock In \emph{Proceedings of the IEEE/CVF Conference on Computer Vision
  and Pattern Recognition}, pages 13339--13348, 2020.

\bibitem[Lin et~al.(2022)Lin, Duan, Zhang, Guo, and Cheng]{Lin_2022_CVPR}
Zheng Lin, Zheng-Peng Duan, Zhao Zhang, Chun-Le Guo, and Ming-Ming Cheng.
\newblock Focuscut: Diving into a focus view in interactive segmentation.
\newblock In \emph{Proceedings of the IEEE/CVF Conference on Computer Vision
  and Pattern Recognition (CVPR)}, pages 2637--2646, June 2022.

\bibitem[Ling et~al.(2019)Ling, Gao, Kar, Chen, and Fidler]{ling2019fast}
Huan Ling, Jun Gao, Amlan Kar, Wenzheng Chen, and Sanja Fidler.
\newblock Fast interactive object annotation with curve-gcn.
\newblock In \emph{Proceedings of the IEEE/CVF Conference on Computer Vision
  and Pattern Recognition}, pages 5257--5266, 2019.

\bibitem[Liu et~al.(2022{\natexlab{a}})Liu, Xu, Bertasius, and
  Niethammer]{liu2022simpleclick}
Qin Liu, Zhenlin Xu, Gedas Bertasius, and Marc Niethammer.
\newblock Simpleclick: Interactive image segmentation with simple vision
  transformers.
\newblock \emph{arXiv preprint arXiv:2210.11006}, 2022{\natexlab{a}}.

\bibitem[Liu et~al.(2022{\natexlab{b}})Liu, Zheng, Planche, Karanam, Chen,
  Niethammer, and Wu]{liu2022pseudoclick}
Qin Liu, Meng Zheng, Benjamin Planche, Srikrishna Karanam, Terrence Chen, Marc
  Niethammer, and Ziyan Wu.
\newblock Pseudoclick: Interactive image segmentation with click imitation.
\newblock In \emph{European Conference on Computer Vision}, pages 728--745.
  Springer, 2022{\natexlab{b}}.

\bibitem[Majumder and Yao(2019)]{majumder2019content}
Soumajit Majumder and Angela Yao.
\newblock Content-aware multi-level guidance for interactive instance
  segmentation.
\newblock In \emph{Proceedings of the IEEE/CVF Conference on Computer Vision
  and Pattern Recognition}, pages 11602--11611, 2019.

\bibitem[Rother et~al.(2004)Rother, Kolmogorov, and Blake]{rother2004grabcut}
Carsten Rother, Vladimir Kolmogorov, and Andrew Blake.
\newblock " grabcut" interactive foreground extraction using iterated graph
  cuts.
\newblock \emph{ACM transactions on graphics (TOG)}, 23\penalty0 (3):\penalty0
  309--314, 2004.

\bibitem[Scarselli et~al.(2008)Scarselli, Gori, Tsoi, Hagenbuchner, and
  Monfardini]{scarselli2008graph}
Franco Scarselli, Marco Gori, Ah~Chung Tsoi, Markus Hagenbuchner, and Gabriele
  Monfardini.
\newblock The graph neural network model.
\newblock \emph{IEEE transactions on neural networks}, 20\penalty0
  (1):\penalty0 61--80, 2008.

\bibitem[Sofiiuk et~al.(2019)Sofiiuk, Barinova, and
  Konushin]{sofiiuk2019adaptis}
Konstantin Sofiiuk, Olga Barinova, and Anton Konushin.
\newblock Adaptis: Adaptive instance selection network.
\newblock In \emph{Proceedings of the IEEE/CVF International Conference on
  Computer Vision}, pages 7355--7363, 2019.

\bibitem[Sofiiuk et~al.(2020)Sofiiuk, Petrov, Barinova, and
  Konushin]{sofiiuk2020f}
Konstantin Sofiiuk, Ilia Petrov, Olga Barinova, and Anton Konushin.
\newblock f-brs: Rethinking backpropagating refinement for interactive
  segmentation.
\newblock In \emph{Proceedings of the IEEE/CVF Conference on Computer Vision
  and Pattern Recognition}, pages 8623--8632, 2020.

\bibitem[Sofiiuk et~al.(2021)Sofiiuk, Petrov, and
  Konushin]{sofiiuk2021reviving}
Konstantin Sofiiuk, Ilia~A Petrov, and Anton Konushin.
\newblock Reviving iterative training with mask guidance for interactive
  segmentation.
\newblock \emph{arXiv preprint arXiv:2102.06583}, 2021.

\bibitem[Wang et~al.(2018{\natexlab{a}})Wang, Zuluaga, Li, Pratt, Patel,
  Aertsen, Doel, David, Deprest, Ourselin, et~al.]{wang2018deepigeos}
Guotai Wang, Maria~A Zuluaga, Wenqi Li, Rosalind Pratt, Premal~A Patel, Michael
  Aertsen, Tom Doel, Anna~L David, Jan Deprest, S{\'e}bastien Ourselin, et~al.
\newblock Deepigeos: a deep interactive geodesic framework for medical image
  segmentation.
\newblock \emph{IEEE transactions on pattern analysis and machine
  intelligence}, 41\penalty0 (7):\penalty0 1559--1572, 2018{\natexlab{a}}.

\bibitem[Wang et~al.(2018{\natexlab{b}})Wang, Girshick, Gupta, and
  He]{wang2018non}
Xiaolong Wang, Ross Girshick, Abhinav Gupta, and Kaiming He.
\newblock Non-local neural networks.
\newblock In \emph{Proceedings of the IEEE conference on computer vision and
  pattern recognition}, pages 7794--7803, 2018{\natexlab{b}}.

\bibitem[Wu et~al.(2014)Wu, Zhao, Zhu, Luo, and Tu]{wu2014milcut}
Jiajun Wu, Yibiao Zhao, Jun-Yan Zhu, Siwei Luo, and Zhuowen Tu.
\newblock Milcut: A sweeping line multiple instance learning paradigm for
  interactive image segmentation.
\newblock In \emph{Proceedings of the IEEE Conference on Computer Vision and
  Pattern Recognition}, pages 256--263, 2014.

\bibitem[Wu et~al.(2020)Wu, Pan, Chen, Long, Zhang, and
  Philip]{wu2020comprehensive}
Zonghan Wu, Shirui Pan, Fengwen Chen, Guodong Long, Chengqi Zhang, and S~Yu
  Philip.
\newblock A comprehensive survey on graph neural networks.
\newblock \emph{IEEE transactions on neural networks and learning systems},
  32\penalty0 (1):\penalty0 4--24, 2020.

\bibitem[Xu et~al.(2016)Xu, Price, Cohen, Yang, and Huang]{xu2016deep}
Ning Xu, Brian Price, Scott Cohen, Jimei Yang, and Thomas~S Huang.
\newblock Deep interactive object selection.
\newblock In \emph{Proceedings of the IEEE Conference on Computer Vision and
  Pattern Recognition}, pages 373--381, 2016.

\bibitem[Zhang et~al.(2020)Zhang, Liew, Wei, Wei, and
  Zhao]{zhang2020interactive}
Shiyin Zhang, Jun~Hao Liew, Yunchao Wei, Shikui Wei, and Yao Zhao.
\newblock Interactive object segmentation with inside-outside guidance.
\newblock In \emph{Proceedings of the IEEE/CVF Conference on Computer Vision
  and Pattern Recognition}, pages 12234--12244, 2020.

\end{thebibliography}
